# Active Diagnosis via AUC Maximization: An Efficient Approach for Multiple Fault Identification in Large Scale, Noisy Networks

**Gowtham Bellala, Jason Stanley, Clayton Scott**
Electrical Engineering and Computer Science
University of Michigan, Ann Arbor

**Suresh K. Bhavnani**
Institute for Translational Sciences
University of Texas Medical Branch, Galveston

## Abstract

The problem of active diagnosis arises in several applications such as disease diagnosis, and fault diagnosis in computer networks, where the goal is to rapidly identify the binary states of a set of objects (e.g., faulty or working) by sequentially selecting, and observing, (noisy) responses to binary valued queries. Current algorithms in this area rely on loopy belief propagation for active query selection. These algorithms have an exponential time complexity, making them slow and even intractable in large networks. We propose a rank-based greedy algorithm that sequentially chooses queries such that the area under the ROC curve of the rank-based output is maximized. The AUC criterion allows us to make a simplifying assumption that significantly reduces the complexity of active query selection (from exponential to near quadratic), with little or no compromise on the performance quality.

## 1 Introduction

The problem of diagnosis appears in various applications such as medical diagnosis (Heckerman, 1990; Jaakkola and Jordan, 1999), fault diagnosis in nuclear plants (Santoso et al., 1999), computer networks (Rish et al., 2005; Zheng et al., 2005), power-delivery systems (Yongli et al., 2006), decoding of messages sent through a noisy channel, etc. In these problems, the goal is to identify the binary states $\mathbf{X} = (X_1, \cdots, X_M)$ of $M$ different objects based on the binary outcomes $\mathbf{Z} = (Z_1, \cdots, Z_N)$ of $N$ distinct queries/tests, where the query responses are noisy. For example, in the problem of medical diagnosis, the goal is to identify the presence/absence of a set of diseases based on the outcomes of medical tests. Similarly, in a fault diagnosis problem, the goal is to identify the state (faulty/working) of each component based on alarm/probe responses (In the rest of this paper, we will refer to an object with state 1 as a fault). In recent years, this problem has been formulated as an inference problem on a Bayesian network, with the goal of assigning most likely states to unobserved object nodes based on the outcome of the query nodes.

An important issue in diagnosis is the trade-off between the cost of querying (uncovering the value of some $Z_j$) and the achieved accuracy of diagnosis. It is often too expensive, time consuming or even impossible to get responses to all the queries. In this paper, we study the problem of *active diagnosis*, where the queries are selected sequentially to maximize the accuracy of diagnosis while minimizing the cost of querying.

Zheng et al. (2005) proposed the use of reduction in conditional entropy (equivalently, mutual information) as a measure to select the most informative subset of queries. They proposed an algorithm that uses the loopy belief propagation (BP) framework to select queries sequentially based on the gain in mutual information, given the observed responses to past queries. This algorithm, which they refer to as BPEA, requires one run of BP for each query selection. Finally, the objects are assigned the most likely states based on the outcome of the selected queries, using a MAP (maximum *a posteriori*) inference algorithm. Refer to Section 3.1 for more details.

However, there are two limitations with this approach. First, the MAP estimate does not equal the true state vector $\mathbf{X}$, either due to noise in the observed query responses or due to suboptimal convergence of the MAP inference algorithm. This leads to false alarm and miss rates that may not be tolerable for a given application.

The second issue is that BPEA does not scale to large networks, because the complexity of computing the approximate value of conditional entropy grows ex-ponentially in the maximum degree of the underlying Bayesian network (see Section 3.1 for details). As we show in Section 5, it becomes intractable even in networks with a few thousand objects. In addition, since this approach relies on belief propagation (BP), it may suffer from the limitations of BP such as slow convergence or oscillation of the algorithm, especially when the prior fault probability is small (Murphy et al., 1999). As we discuss below, the prior fault probability is often very low in real-world diagnosis problems.

In this paper, we propose to address these two limitations by adopting an AUC (Area under the ROC curve) criterion for query selection. To address the first limitation, we propose to output a ranked list of objects rather than their most likely states, where the ranking is based on their posterior fault probability. Given such a ranked list, the object states can be estimated by choosing a threshold $t$, where the top $t$ objects in the ranked list are declared as faults (i.e., state 1) and the remaining as 0. Varying $t$ gives a receiver operating characteristic (ROC) curve. We show how to select queries sequentially by maximizing AUC.

The rank-based approach is motivated by the fact that in many applications there is a domain expert who makes the final decision on the objects' states. Such a ranking can be useful to a domain expert who will use domain expertise and other sources of information to choose a threshold $t$ that may lead to a permissible value of false alarm and miss rates for a given application.

To address the second limitation, we circumvent the use of BP in the query selection stage by making the simplifying assumption of a single fault, i.e., the state of only one object can be equal to 1. To be clear, we still intend to apply our algorithm when multiple faults are present; the single fault assumption is used in the design of the algorithm. This assumption is reasonable because the prior fault probability is quite low in many applications. For example, in the problem of fault diagnosis in computer networks, the prior probability of a router failing in any given hour is on the order of $10^{-6}$ (Kandula et al., 2005). Similarly, in the disease diagnosis problem of QMR-DT, the prior probability of a disease being "present" is typically on the order of $10^{-3}$ (Murphy et al., 1999).

We show that the AUC criterion can be optimized efficiently under the single-fault assumption. While other criteria such as mutual information can also be optimized efficiently under this assumption, we show that AUC is much more robust to violations of the single fault assumption, which are bound to happen in practice. We demonstrate through experiments on computer networks that the proposed query selection cri-terion can achieve performance close to that of BPEA *in a multi-fault setting*, while having a computational complexity that is orders less than that of BPEA. Thus, it is a fast and a reliable substitute for BPEA in large scale diagnosis problems.

## 2 Data Model

A diagnosis problem is often represented by a bipartite diagnosis graph (BDG) between a set of $M$ different objects, and a set of $N$ distinct queries, with edges between the two entities. These edges represent the relation or the interactions between the two entities. For example, in a fault diagnosis problem, the objects correspond to components and queries to alarms where an edge determines if a particular component-alarm pair is connected. Similarly, in a disease diagnosis problem, objects may correspond to diseases and queries to symptoms where an edge determines if a particular symptom is exhibited by a disease. Figure 1 demonstrates a toy bipartite diagnosis graph.

We denote the state of each object (e.g., presence/absence of a disease) with a binary random variable $X_i$ and the state of each query (i.e., the observed response to a query) by a binary random variable $Z_j$. Then, $\mathbf{X} = (X_1, \cdots, X_M)$ is a binary random vector denoting the states of all the objects, and $\mathbf{Z} = (Z_1, \cdots, Z_N)$ is a binary random vector denoting the responses to all the queries, where $\mathbf{x} \in \{0,1\}^M$ and $\mathbf{z} \in \{0,1\}^N$ correspond to realizations of $\mathbf{X}$ and $\mathbf{Z}$, respectively.

In addition, for any subset of queries $\mathcal{A} \subseteq \{1, \cdots, N\}$, we denote by $\mathbf{Z}_\mathcal{A}$ the random variables associated with those queries, e.g., if $\mathcal{A} = \{1,4,7\}$, then $\mathbf{Z}_\mathcal{A} = (Z_1, Z_4, Z_7)$. Also, for any query $j$, let $\mathbf{pa}_j$ denote the objects that are connected to it in the BDG. Then, $\mathbf{X}_{\mathbf{pa}_j}$ denotes the states of all the objects connected to query $j$, e.g., for query 2 in Figure 1, $\mathbf{X}_{\mathbf{pa}_2} = (X_2, X_3)$.

We need to specify the joint distribution of $(\mathbf{X}, \mathbf{Z})$, and more generally $(\mathbf{X}, \mathbf{Z}_\mathcal{A})$ for any $\mathcal{A}$, which can be defined in terms of a prior probability distribution on $\mathbf{X}$ and a conditional distribution on $\mathbf{Z}_\mathcal{A}$ given $\mathbf{X}$. To define the prior probability distribution on $\mathbf{X}$, we make the standard assumption that the object states are marginally independent, i.e., $\Pr(\mathbf{X} = \mathbf{x}) = \prod_{i=1}^{M} \Pr(X_i = x_i)$. Similarly, to define the conditional distribution on $\mathbf{Z}_\mathcal{A}$ given $\mathbf{X}$, we make the standard assumption that the observed responses to queries are conditionally independent given the states of the objects connected to them, i.e.,

$$\Pr(\mathbf{Z}_\mathcal{A} = \mathbf{z}_\mathcal{A} | \mathbf{X} = \mathbf{x}) = \prod_{j \in \mathcal{A}} \Pr(Z_j = z_j | \mathbf{x}_{\mathbf{pa}_j}).$$

These assumptions hold reasonably well in many prac-tical applications. For example, in a fault diagnosis problem, it can be reasonable to assume that the components fail independently and that the alarm responses are conditionally independent given the states of the components they are connected to. These dependencies can be encoded by a Bayesian network as shown in Figure 1.

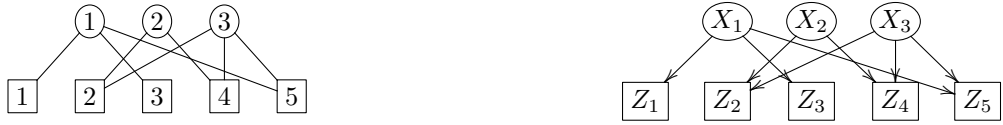

Figure 1: (left) A toy bipartite diagnosis graph (BDG) where the circled nodes denote the objects and the square nodes denote queries. (right) A Bayesian Network corresponding to the given BDG.

In the ideal case when there is no noise, the observed response $Z_j$ to query $j$ is deterministic given the binary states of the objects in $\mathbf{pa}_j$. Specifically, it is given by the OR operation of the binary variables in $\mathbf{X}_{\mathbf{pa}_j}$, i.e., $Z_j = 1 \iff \exists\, i \in \mathbf{pa}_j$ s.t. $X_i = 1$. In general, it is a noisy OR operation where the conditional distribution of $Z_j$ given $\mathbf{x}_{\mathbf{pa}_j}$ can be defined using standard noise models such as the Y-model (Le and Hadjicostis, 2007) or the QMR-DT model (Pearl, 1988).

We derive the AUC based active diagnosis algorithm under this general probability model, and in Section 5, we demonstrate the performance of the proposed algorithm in the problem of fault diagnosis in computer networks under the QMR-DT noise model, where

$$\Pr(X_i = x) := (\alpha_i)^x (1 - \alpha_i)^{1-x}, \text{ and}$$
$$\Pr(Z_j = 0 | \mathbf{x}_{\mathbf{pa}_j}) := \rho_{0j} \prod_{k \in \mathbf{pa}_j} \rho_{kj}^{x_k}.$$

Here, $\alpha_i$ is the prior fault probability, $\rho_{kj}$ and $(1 - \rho_{0j})$ are the so-called inhibition and leak probabilities, respectively.

## 3 Active Diagnosis

The approach in active diagnosis is to maximize some function $f(\mathbf{z}_\mathcal{A})$, which denotes the quality of the estimate of $\mathbf{X}$, subject to a constraint on the number of queries made, i.e.,

$$\max_{\mathcal{A} \subseteq \{1, \cdots, N\}} f(\mathbf{z}_\mathcal{A})$$
$$\text{s.t.} \quad |\mathcal{A}| \leq k.$$

In general, finding an optimal solution to this problem is NP-hard (Rish et al., 2005). Instead, the queries can be chosen sequentially by greedily maximizing the quality function, given the observed responses to the past queries, i.e.,

$$j^* := \underset{j \notin \mathcal{A}}{\operatorname{argmax}}\ \mathbb{E}_{Z_j}[f(\mathbf{z}_\mathcal{A} \cup Z_j) - f(\mathbf{z}_\mathcal{A}) | \mathbf{Z}_\mathcal{A} = \mathbf{z}_\mathcal{A}] \quad (1)$$

where $\mathbf{z}_\mathcal{A} \cup Z_j$ denotes the observed responses to queries in $\mathcal{A} \cup \{j\}$.

### 3.1 Entropy-based Active Query Selection

Zheng et al. (2005) proposed the use of the reduction in conditional entropy (equivalently, mutual information), as a quality function, namely $f(\mathbf{z}_\mathcal{A}) = H(\mathbf{X}) - H(\mathbf{X}|\mathbf{z}_\mathcal{A})$. Here, given the observed responses $\mathbf{z}_\mathcal{A}$ to previously selected queries $\mathcal{A}$, the next query is chosen to be

$$j^* = \underset{j \notin \mathcal{A}}{\operatorname{argmin}} \sum_{z=0,1} \Pr(Z_j = z | \mathbf{z}_\mathcal{A}) H(\mathbf{X} | \mathbf{z}_\mathcal{A}, z) \quad (2)$$

where the conditional entropy is given by

$$H(\mathbf{X}|\mathbf{z}_\mathcal{A}, z) = -\sum_{\mathbf{x} \in \{0,1\}^M} \Pr(\mathbf{x}|\mathbf{z}_\mathcal{A}, z) \log_2 \Pr(\mathbf{x}|\mathbf{z}_\mathcal{A}, z).$$

Note that direct computation of the above expression is intractable. However, under the independence assumptions of Section 2, the conditional entropy can be simplified such that the query selection criterion in (2) is reduced to

$$\underset{j \notin \mathcal{A}}{\operatorname{argmin}} \left[ -\sum_{\mathbf{x}_{\mathbf{pa}_j}, z} \Pr(\mathbf{x}_{\mathbf{pa}_j}, z | \mathbf{z}_\mathcal{A}) \log_2 \Pr(Z_j = z | \mathbf{x}_{\mathbf{pa}_j}) \right.$$
$$\left. + \sum_{z=0,1} \Pr(Z_j = z | \mathbf{z}_\mathcal{A}) \log_2 \Pr(Z_j = z | \mathbf{z}_\mathcal{A}) + \text{const} \right].$$

Zheng et al. (2005) proposed an approximation algorithm that uses the BP infrastructure to compute the above expression, which they refer to as belief propagation for entropy approximation (BPEA). This algorithm requires one run of BP for each query selection. After observing responses $\mathbf{z}_\mathcal{A}$ to a set of queries $\mathcal{A}$, the object states are then estimated to be

$$\mathbf{x}^{\text{MAP}} := \underset{\mathbf{x} \in \{0,1\}^M}{\operatorname{argmax}}\ \Pr(\mathbf{X} = \mathbf{x} | \mathbf{z}_\mathcal{A}).$$

However, this approach does not scale to large networks as BPEA involves a term whose computation grows exponentially in the number of parents to a query node. If $m$ denotes the maximum number of parents to any query node, i.e., $m := \max_{j \in \{1, \cdots, N\}} |\mathbf{pa}_j|$, then the computational complexity of choosing a query

| | $X_1$ | $X_2$ | $X_3$ | $X_4$ | $X_5$ |
|---|---|---|---|---|---|
| $\Pr(X_i = 1\|\mathbf{z}_\mathcal{A})$ | 0.3 | 0.15 | 0.35 | 0.15 | 0.05 |

Figure 2: A rank order corresponding to this example is $\mathbf{r} = (3, 1, 2, 4, 5)$.

using BPEA is $O(N2^m)$, thus making it intractable in networks where $m$ is greater than 25.

Recently, Cheng et al. (2010) proposed a speed up to query selection using BPEA by reducing the number of queries to be investigated at each stage. However, the exponential complexity still remains. Moreover, though this objective can be computed efficiently under a single fault assumption, as we explain in Section 4.2, entropy-based query selection under a single fault assumption can perform poorly in a multi-fault setting.

In the next section, we derive a new query selection criterion that sequentially chooses queries such that the area under the ROC curve of the rank-based output is maximized. In Section 4.1, we show that the proposed query selection criterion can be implemented efficiently under a single fault assumption, and in Section 5, we show that the AUC-based query selection can achieve performance close to that of BPEA, even when multiple faults occur, thus making it a viable substitute for BPEA in large scale networks.

## 4 AUC-based Active Query Selection

AUC has been used earlier as a performance criterion in the classification setting with decision tree classifiers (Ferri et al., 2002; Cortes and Mohri, 2003) and boosting (Long and Servedio, 2007), and in the problem of ranking (Ataman et al., 2006), but not in an active diagnosis setting to the best of our knowledge. In all the earlier settings, the AUC of a classifier is estimated using the training data whose binary labels are known. However, in an active diagnosis setting, the object states (binary labels) are neither known nor does there exist any training data. Hence, we propose a simple estimator for the AUC, based on the posterior probabilities of the object states.

Given the observed responses $\mathbf{z}_\mathcal{A}$ to queries in $\mathcal{A}$, let the objects be ranked based on their posterior fault probabilities, i.e., $\Pr(X_i = 1|\mathbf{z}_\mathcal{A})$, where ties involving objects with the same posterior probability are broken randomly. Then, let $\mathbf{r} = (r(1), \cdots, r(M))$ denote the rank order of the objects, where $r(i)$ denotes the index of the $i$th ranked object. For example, a rank order corresponding to the toy example in Figure 2 is $\mathbf{r} = (3, 1, 2, 4, 5)$. Also, note that $\mathbf{r}$ depends on the queries chosen $\mathcal{A}$ and their observed responses $\mathbf{z}_\mathcal{A}$, though it is not explicitly shown in our notation.

Given this ranked list of objects, we get a series of estimators $\{\widehat{\mathbf{x}}^t\}_{t=0}^M$ for the object state vector $\mathbf{X}$, where $\widehat{\mathbf{x}}^t$ corresponds to the estimator which declares the states of the top $t$ objects in the ranked list as 1 and the remaining as 0. For example, $\widehat{\mathbf{x}}^2 = (1, 0, 1, 0, 0)$ for the toy example shown in Figure 2.

These estimators have different false alarm and miss rates. The miss and false alarm rates associated with $\widehat{\mathbf{x}}^t$ are given by

$$\mathrm{MR}_t = \frac{\sum_{\{i:\widehat{x}_i^t=0\}} \mathbf{I}\{X_i = 1\}}{\sum_{i=1}^M \mathbf{I}\{X_i = 1\}} = \frac{\sum_{i=t+1}^M \mathbf{I}\{X_{r(i)} = 1\}}{\sum_{i=1}^M \mathbf{I}\{X_i = 1\}},$$

$$\mathrm{FAR}_t = \frac{\sum_{\{i:\widehat{x}_i^t=1\}} \mathbf{I}\{X_i = 0\}}{\sum_{i=1}^M \mathbf{I}\{X_i = 0\}} = \frac{\sum_{i=1}^t \mathbf{I}\{X_{r(i)} = 0\}}{\sum_{i=1}^M \mathbf{I}\{X_i = 0\}},$$

where $\mathbf{I}\{E\}$ is an indicator function which takes the value 1 when the event $E$ is true, and 0 otherwise.

However, since the true states of the objects are not known, the false alarm and the miss rates should be estimated. Based on the responses $\mathbf{z}_\mathcal{A}$ to queries in $\mathcal{A}$, these two error rates can be estimated as

$$\widehat{\mathrm{MR}}_t(\mathbf{z}_\mathcal{A}) = \frac{\sum_{i=t+1}^M \Pr(X_{r(i)} = 1|\mathbf{z}_\mathcal{A})}{\sum_{i=1}^M \Pr(X_i = 1|\mathbf{z}_\mathcal{A})}, \quad (3a)$$

$$\widehat{\mathrm{FAR}}_t(\mathbf{z}_\mathcal{A}) = \frac{\sum_{i=1}^t \Pr(X_{r(i)} = 0|\mathbf{z}_\mathcal{A})}{\sum_{i=1}^M \Pr(X_i = 0|\mathbf{z}_\mathcal{A})}. \quad (3b)$$

Using these estimates, the ROC curve can then be obtained by varying the threshold $t$ from 0 to $M$ leading to different false alarm and miss rates. For example, $\widehat{\mathbf{x}}^0$ which declares the states of all the objects to be equal to 0, has a false alarm rate of 0 and a miss rate of 1. On the other hand, $\widehat{\mathbf{x}}^M$ which declares the states of all objects as 1, has a false alarm rate of 1 with a miss rate of 0. The other estimators have false alarm and miss rates that span the space between these two extremes.

Finally, the area under this ROC curve can be estimated using a piecewise approximation with either lower rectangles, upper rectangles or a linear approximation (refer Bellala et al. (2011) for more details). For reasons discussed below, upper rectangles work better than the other two options. The AUC estimate based on upper rectangles is given by

$$\mathbf{A}(\mathbf{z}_\mathcal{A}) := \widehat{\mathrm{AUC}} = \sum_{t=0}^{M-1} (1 - \widehat{\mathrm{MR}}_{t+1})(\widehat{\mathrm{FAR}}_{t+1} - \widehat{\mathrm{FAR}}_t)$$

$$= \widehat{\mathrm{FAR}}_M - \widehat{\mathrm{FAR}}_0 - \sum_{t=0}^{M-1} \widehat{\mathrm{MR}}_{t+1}(\widehat{\mathrm{FAR}}_{t+1} - \widehat{\mathrm{FAR}}_t)$$

$$= 1 - \sum_{t=0}^{M-1} \widehat{\mathrm{MR}}_{t+1}(\widehat{\mathrm{FAR}}_{t+1} - \widehat{\mathrm{FAR}}_t),$$

where we dropped the dependence of $\widehat{\text{MR}}_t$ and $\widehat{\text{FAR}}_t$ on $\mathbf{z}_\mathcal{A}$ to avoid cramping. Also, note that $\sum_{t=0}^{M-1} \widehat{\text{MR}}_{t+1}(\widehat{\text{FAR}}_{t+1} - \widehat{\text{FAR}}_t)$ corresponds to an estimate of the area *above* the ROC curve, which we denote by $\overline{\mathbf{A}}(\mathbf{z}_\mathcal{A})$.

Given this quality function, the goal of active diagnosis is to maximize the accuracy of diagnosis given by $\underline{\mathbf{A}}(\mathbf{z}_\mathcal{A})$, subject to a constraint on the number of queries made, i.e.,

$$\max_{\mathcal{A} \subseteq \{1, \cdots, N\}} \underline{\mathbf{A}}(\mathbf{z}_\mathcal{A})$$
$$\text{s.t.} \quad |\mathcal{A}| \leq k.$$

Substituting this quality function in (1), we get the criterion for greedily choosing the next query to be

$$j^* = \operatorname*{argmin}_{j \notin \mathcal{A}} \sum_{z=0,1} \Pr(Z_j = z | \mathbf{z}_\mathcal{A}) \overline{\mathbf{A}}(\mathbf{z}_\mathcal{A} \cup z). \quad (4)$$

Substituting the estimates of miss rate and false alarm rate from (3a) and (3b), we get $\overline{\mathbf{A}}(\mathbf{z}_\mathcal{A})$ to be

$$\frac{\sum_{i=1}^{M-1} \left[ \Pr(X_{r(i)} = 0 | \mathbf{z}_\mathcal{A}) \sum_{j=i+1}^{M} \Pr(X_{r(j)} = 1 | \mathbf{z}_\mathcal{A}) \right]}{\sum_{i=1}^{M} \Pr(X_i = 1 | \mathbf{z}_\mathcal{A}) \sum_{i=1}^{M} \Pr(X_i = 0 | \mathbf{z}_\mathcal{A})}. \quad (5)$$

Note that both the query selection criterion in (4) and the function $\overline{\mathbf{A}}(\mathbf{z}_\mathcal{A})$ in (5) depend only on the posterior probabilities of unobserved nodes given the states of the observed nodes. Since these probabilities can be approximated using BP, the AUC-based active query selection can be performed using BP similar to the entropy-based active query selection.

However, the main focus of this paper is on active diagnosis for large scale networks where query selection using BP is slow and possibly intractable. In the next section, we show that the proposed AUC-based query selection can be performed efficiently under a single fault assumption.

### 4.1 Single Fault Assumption

We now derive the AUC-based query selection criterion under the single fault assumption. Under this assumption, the object state vector $\mathbf{X}$ is restricted to belong to the set $\{\mathbb{I}_1, \cdots, \mathbb{I}_M\}$ in the query selection stage, where $\mathbb{I}_i$ is a binary vector whose $i$th element is 1 and the remaining elements are 0. This reduction in the state space of the object vector allows for query selection to be performed efficiently without the need for BP.

More specifically, the posterior probabilities required to choose queries sequentially in (4) can be computed as follows. Using the conditional independence assumption, $\Pr(Z = z | \mathbf{z}_\mathcal{A})$ can be computed as

$$\Pr(Z = z | \mathbf{z}_\mathcal{A}) = \sum_{i=1}^{M} \Pr(Z = z | \mathbf{X} = \mathbb{I}_i) \Pr(\mathbf{X} = \mathbb{I}_i | \mathbf{z}_\mathcal{A}),$$

where $\Pr(\mathbf{X} = \mathbb{I}_i | \mathbf{z}_\mathcal{A})$ can be computed directly as

$$\Pr(\mathbf{X} = \mathbb{I}_i | \mathbf{z}_\mathcal{A}) = \frac{\Pr(\mathbf{X} = \mathbb{I}_i) \Pr(\mathbf{Z}_\mathcal{A} = \mathbf{z}_\mathcal{A} | \mathbf{X} = \mathbb{I}_i)}{\sum_{j=1}^{M} \Pr(\mathbf{X} = \mathbb{I}_j) \Pr(\mathbf{Z}_\mathcal{A} = \mathbf{z}_\mathcal{A} | \mathbf{X} = \mathbb{I}_j)}$$

with $\Pr(\mathbf{Z}_\mathcal{A} = \mathbf{z}_\mathcal{A} | \mathbf{X} = \mathbb{I}_i) = \prod_{k \in \mathcal{A}} \Pr(Z_k = z_k | \mathbf{X} = \mathbb{I}_i)$. Also, note that under a single fault assumption,

$$\sum_{i=1}^{M} \Pr(X_i = 1 | \mathbf{z}_\mathcal{A}) = \sum_{i=1}^{M} \Pr(\mathbf{X} = \mathbb{I}_i | \mathbf{z}_\mathcal{A}) = 1, \quad (6)$$

$$\sum_{i=1}^{M} \Pr(X_i = 0 | \mathbf{z}_\mathcal{A}) = \sum_{i=1}^{M} 1 - \Pr(X_i = 1 | \mathbf{z}_\mathcal{A}) = M - 1.$$

Hence, the estimate of the area above the ROC curve $\overline{\mathbf{A}}(\mathbf{z}_\mathcal{A})$ in (5) reduces to

$$\frac{\sum_{i=1}^{M-1} \left[ \Pr(X_{r(i)} = 0 | \mathbf{z}_\mathcal{A}) \sum_{j=i+1}^{M} \Pr(X_{r(j)} = 1 | \mathbf{z}_\mathcal{A}) \right]}{M - 1}. \quad (7)$$

The following result gives an efficient way to compute $\overline{\mathbf{A}}(\mathbf{z}_\mathcal{A})$.

**Proposition 1.** *The estimate of the area above the ROC curve $\overline{\mathbf{A}}(\mathbf{z}_\mathcal{A})$ in (5) can be equivalently expressed as*

$$\frac{1}{2} + \frac{\sum_{i=1}^{M}(2i - M - 2)\Pr(X_{r(i)} = 1 | \mathbf{z}_\mathcal{A}) + \Pr^2(X_i = 1 | \mathbf{z}_\mathcal{A})}{2 \sum_{i=1}^{M} \Pr(X_i = 1 | \mathbf{z}_\mathcal{A}) \sum_{i=1}^{M} \Pr(X_i = 0 | \mathbf{z}_\mathcal{A})}$$

From this result, given a ranked list of the objects, the complexity of computing $\overline{\mathbf{A}}(\mathbf{z}_\mathcal{A})$ is $O(M)$, i.e., the complexity of computing $\overline{\mathbf{A}}(\mathbf{z}_\mathcal{A})$ is dominated by the complexity of sorting, $O(M \log M)$. Hence, the computational complexity of choosing a query at each stage using the AUC-based criterion under a single fault assumption is $O(NM \log M)$.

In addition, as we show in Theorem 1 below, AUC estimated using lower rectangles or a linear approximation is adaptive monotone (Golovin and Krause, 2010), i.e., the accuracy of diagnosis given by $\underline{\mathbf{A}}(\mathbf{Z}_\mathcal{A})$ is guaranteed to increase by acquiring more query information (equivalently, the area above the ROC curve given by $\overline{\mathbf{A}}(\mathbf{Z}_\mathcal{A})$ is guaranteed to decrease by acquiring more query information).

**Theorem 1.** *Under the single fault assumption, the quality function $\underline{\mathbf{A}}(\mathbf{Z}_{\mathcal{A}})$ estimated using either lower rectangles or a linear approximation, is adaptive monotone, i.e., $\forall \mathcal{A}' \subseteq \mathcal{A}$*

$$\underline{\mathbf{A}}(\mathbf{Z}_{\mathcal{A}'}) \leq \underline{\mathbf{A}}(\mathbf{Z}_{\mathcal{A}})$$

Refer Bellala et al. (2011) for proofs of Proposition 1 and Theorem 1.

### 4.2 Comparison with Single-Fault Entropy

As mentioned earlier, the reduction in conditional entropy can also be computed efficiently under a single fault assumption. However, entropy-based query selection under a single fault assumption performs poorly in a multi-fault setting. We will now provide an intuitive explanation for this phenomenon. Our argument relies on the following result, whose proof can be found in (Bellala et al., 2011).

**Proposition 2.** *Under the single fault assumption, along with the conditional independence assumption of Section 2, the entropy-based query selection criterion in (2) reduces to*

$$j^* := \operatorname*{argmin}_{j \notin \mathcal{A}} \sum_{i=1}^{M} \Pr(X_i = 1|\mathbf{z}_{\mathcal{A}}) H\Big(\Pr(Z_j = 0|X_i = 1)\Big)$$
$$- H\Big(\Pr(Z_j = 0|\mathbf{z}_{\mathcal{A}})\Big) \quad (8)$$

*where $H(p) := -p \log_2 p - (1-p) \log_2(1-p)$ denotes the binary entropy function.*

As noted in (6), under a single fault assumption, the posterior fault probabilities are constrained to sum to 1. Hence, objects with high posterior fault probability decrease the posterior fault probabilities of the remaining objects. Given this scenario, note from (8) in Proposition 2, that both the terms in this query selection criterion are highly dominated by the object(s) with high posterior fault probabilities (even the second term, since $\Pr(Z_j = 0|\mathbf{z}_{\mathcal{A}}) = \sum_{i=1}^{M} \Pr(X_i = 1|\mathbf{z}_{\mathcal{A}}) \Pr(Z_j = 0|X_i = 1)$). Hence, at any given instance, the query chosen according to this criterion is highly biased towards objects that already have a high posterior fault probability. This could lead to a poor choice of queries as the objects with high posterior fault probability need not have their true states as 1, especially in the initial stages.

On the other hand, the AUC-based criterion under single fault assumption in (7) chooses queries at each stage by taking into account its effect on all the objects, leading to a more balanced and informative choice of queries. This can be observed from its expression in (7), by re-writing it as

$$\frac{\sum_{i=2}^{M} \Pr(X_{r(i)} = 1|\mathbf{z}_{\mathcal{A}}) \sum_{j=1}^{i-1} \Pr(X_{r(j)} = 0|\mathbf{z}_{\mathcal{A}})}{M-1},$$

where the object with the least posterior fault probability, i.e., $X_{r(M)}$, is assigned the maximum weight of $\sum_{j=1}^{M-1} \Pr(X_{r(j)} = 0|\mathbf{z}_{\mathcal{A}})$, with monotonically decreasing weights as the posterior fault probability of the objects increases. This forces to choose a query that takes in to consideration the effect on all the objects.

Though AUC approximated using either lower rectangles or a linear approximation are similarly robust to objects with high posterior fault probabilities, for reasons explained in detail in Bellala et al. (2011), AUC approximated using upper rectangles is a better choice.

## 5 Application: Fault Diagnosis in Computer Networks

In this application, the goal is to monitor a system of networked computers for faults, where each computer can be associated with a binary random variable $X_i$ (0 for working and 1 for faulty). It is not possible to test each individual computer directly in a large network. Hence, a common solution is to test a subset of computers with a single *test probe* $Z_j$, where a probe can be as simple as a ping request or more sophisticated such as an e-mail message or a webpage-access request. Thus, there is a bipartite diagnosis graph with each query (probe) connected to all the objects (computers) it passes through. In these networks, certain computers are designated as *probe stations*, which are instrumented to send out *probes* to test the response of the networked elements. However, the available set of probes $\mathbf{Z}$ is often very large, and hence it is desired to minimize the number of probes required to identify the faulty computers. Refer Rish et al. (2005) for further details.

We compare the performance of the proposed AUC-based active query selection under single fault assumption (AUC+SF) with BPEA and entropy-based active query selection under single fault assumption (Entropy+SF), on 1 synthetic dataset and 2 computer networks. Unlike Zheng et al. (2005) and Cheng et al. (2010) who only considered networks of size up to 500 components and 580 probes, here we also consider a large scale network.

The first dataset is a random bipartite diagnosis graph (Guillaume and Latapy, 2004) generated using the standard Preferential Attachment (PA) random graph model. The second and the third datasets are network topologies built using the BRITE (Medina et al., 2001) and the INET (Winick and Jamin, 2002) generators, which simulate an Internet-like topology at the Autonomous Systems level. To generate a BDG of components and probes from these topologies, we used the approach described by Rish et al. (2005) and Zheng et al. (2005).

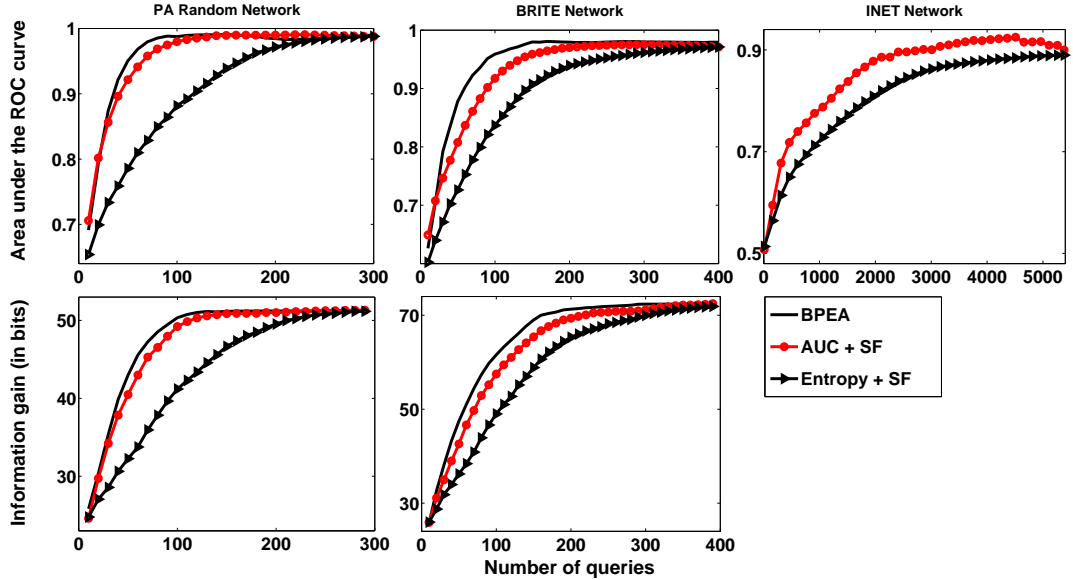

Figure 3: Demonstrates the competitive performance of the AUC-based query selection under single fault assumption to that of BPEA, while having a computational complexity that is orders less (near quadratic vs. exponential complexity of BPEA). On INET, we only compare AUC+SF with Entropy+SF as BPEA becomes slow and intractable.

For the random graph model considered, we generated a random BDG consisting of 300 objects and 300 queries. We generated a BRITE network consisting of 300 components and around 400 probes, and an INET network consisting of 4000 components and 5380 probes. We consider the QMR-DT noise model described in Section 2. We compare the 3 query selection criteria under 2 performance measures, AUC and Information gain.

Figure 3 compares their performance as a function of the number of queries inputted. To compute the area under the ROC curve, we rank the objects based on their posterior fault probabilities that are computed using a single-fault assumption. Alternatively, note that these posterior probabilities could be computed using BP for the PA and BRITE networks (BP is slow and intractable on the INET). For performance of the three query selection criteria under AUC computed with BP based rankings, refer Bellala et al. (2011).

On the other hand, the reduction in conditional entropy is computed approximately using BPEA. We used the inference engines in the libDAI (Mooij, 2010) package for implementing BPEA and BP. However, BPEA (and BP) became slow and intractable on the INET, with BP often not converging and resulting in oscillations. Hence, on this network, we only compare the performance of AUC+SF and Entropy+SF based on the AUC criterion which is computed based on rankings obtained from posterior probabilities under a single-fault assumption.

The results in this figure correspond to a prior fault probability value of 0.03, with the leak and inhibition probabilities at 0.05[1]. Each curve in this figure is averaged over 200 random realizations, where each random realization corresponds to a random state of $\mathbf{X}$ and random generation of the noisy query responses. For the PA and BRITE models, the results were observed to be consistent across different realizations of the underlying bipartite network. For INET, we considered only one network with 25 probe stations.

Note from this figure that AUC+SF invariably performs better than Entropy+SF, and comparable to BPEA. We observed similar comparable performance of AUC+SF to that of BPEA, for different values of leak and inhibition probabilities, and other low values of prior fault probabilities (Bellala et al., 2011). In addition, note from Figure 4 that the time complexity of selecting a query grows exponentially for BPEA, whereas for AUC+SF, it grows near quadratically ($O(NM \log M)$) with the time taken to select a probe being less than 2 seconds even in networks with

---

[1] Refer Bellala et al. (2011) for results on other values of prior, leak and inhibition probabilities

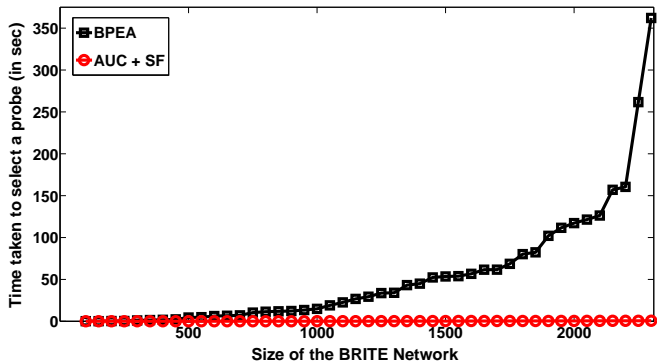

Figure 4: Comparison of time complexity of selecting a query using BPEA and AUC+SF.

2000 components.

These experiments demonstrate the competitive performance of AUC-based query selection under single fault assumption to that of BPEA, besides having a computational complexity that is orders less than that of BPEA, demonstrating its potential as a fast and a reliable substitute for BPEA under low prior, in large scale diagnosis problems.

## 6 Conclusions

We study the problem of active diagnosis for multiple fault identification in large scale, noisy networks. Noting that active query selection algorithms such as BPEA that rely on belief propagation are intractable in large networks, we propose to make the simplifying assumption of a single fault in the query selection stage. Under this assumption, several query selection criterion can be implemented efficiently. However, we note that traditional approaches such as Information gain based query selection under a single fault assumption performs poorly in a multiple fault setting. Hence, we propose a new query selection criterion, where the queries are selected sequentially such that the area under the ROC curve (AUC) of a rank-based output is maximized. We demonstrate the competitive performance of the proposed algorithm to BPEA in the context of fault diagnosis in computer networks. The competitive performance of the proposed algorithm, while having a computational complexity that is orders less than that of BPEA (near quadratic vs. the exponential complexity of BPEA), makes it a fast and a reliable substitute for BPEA in large scale diagnosis problems.

**Acknowledgments**

This work was supported in part by NSF Awards No. 0830490, 0953135 and CDC/NIOSH Grant No. R21 OH009441-01A2.